# An Estimation of Distribution Algorithm with Intelligent Local Search for Rule-based Nurse Rostering

Uwe Aickelin, Edmund K. Burke and Jingpeng Li*
{uxa, ekb, jpl}@cs.nott.ac.uk
School of Computer Science and IT
The University of Nottingham
Nottingham, NG8 1BB
United Kingdom

* Corresponding author. Authors are in alphabetical order.

**Abstract-** This paper proposes a new memetic evolutionary algorithm to achieve explicit learning in rule-based nurse rostering, which involves applying a set of heuristic rules for each nurse's assignment. The main framework of the algorithm is an estimation of distribution algorithm, in which an ant-miner methodology improves the individual solutions produced in each generation. Unlike our previous work (where learning is implicit), the learning in the memetic estimation of distribution algorithm is explicit, i.e. we are able to identify building blocks directly. The overall approach learns by building a probabilistic model, i.e. an estimation of the probability distribution of individual nurse-rule pairs that are used to construct schedules. The local search processor (i.e. the ant-miner) reinforces nurse-rule pairs that receive higher rewards. A challenging real world nurse rostering problem is used as the test problem. Computational results show that the proposed approach outperforms most existing approaches. It is suggested that the learning methodologies suggested in this paper may be applied to other scheduling problems where schedules are built systematically according to specific rules.

## 1 Introduction

Nurse rostering problems have been extensively studied over the past three decades (Burke et al, 2004; Cheang, 2003). Most of these problems are extremely difficult and have been regarded as being more complex than the travelling salesman problem (Tien and Kamiyama, 1982). Early research (e.g. Warner and Prawda, 1972) concentrated on the development of mathematical programming models. To reduce computational complexity, these early researchers had to restrict the problem dimensions and consider a relatively low number of constraints in their models, resulting in methodologies that are too simple to be applied in most modern real world hospital situations. However, modern mathematical programming approaches have been shown to be effective for large problems when combined with heuristic methodologies (e.g. Bard and Purnomo, 2007; Dowsland and Thompson, 2000).

Over the years, there have been many other attempts to solve nurse rostering problems within reasonable time. Artificial intelligence approaches such as constraint programming (e.g. Meyer auf'm Hofe, 2001) and knowledge based systems (e.g. Beddoe and Petrovic, 2006) have been investigated with some success. Since the 1990's, meta-heuristics have attracted the most attention in tackling this problem. Genetic Algorithms (GAs) (e.g. see Sastry et al, 2005) form an important class of meta-heuristics, and have been extensively applied in nurse rostering (e.g. Aickelin and Dowsland, 2000; Aickelin and Dowsland, 2004; Burke et al, 2001; Easton and

Mansour, 1999; Kawanaka et al, 2001). A number of attempts have also been made by using other meta-heuristics, such as simulated annealing (e.g. Brusco and Jacobs, 1995) and tabu search (e.g. Burke et al, 1999; Ikegami and Niwa, 2003).

Most existing GA-based approaches for various nurse rostering problems are direct approaches, in which the individuals in the population represent direct encodings of solutions. However, these problems can also be solved indirectly (Aickelin and Dowsland, 2004). A similar in direct approach has been employed on another personnel rostering problem – driver scheduling (Li and Kwan, 2003; Li and Kwan, 2005). The idea is that we first transform the original problem into another rule-based problem in which we define a set of heuristic rules for the nurses' assignment, and then solve it by using *building* heuristics to construct schedules step by step based on these predefined rules (Aickelin, 2002).

In previous indirect approaches, learning was implicit and restricted to the efficient adjustment of weights for the set of rules that is used to build the schedules (Aickelin and Dowsland, 2004; Li and Kwan, 2003). The major limitation for this type of learning is that once the best weight combination is found, the rules used in the build process are fixed at each iteration. One of the motivations for this paper is based upon the observation that when finishing a chess game, a long sequence of interactive moves with a particular goal in mind is required. An intelligent schedule-building process needs to build its solution in a similar way. Using fixed rules for each move is not consistent with the human learning process. A human scheduler normally builds a schedule systematically by following a set of rules. After much practice, the scheduler gradually masters the knowledge of which solution parts go well with others. He/she can identify good parts and is aware of the level of solution quality even if the scheduling process is not yet completed. The skilled scheduler has the ability to finish a schedule by using flexible, rather than fixed, rules. To establish an explicit learning mechanism from past solutions, two more human-like scheduling algorithms, namely a cutting-edge Bayesian optimization algorithm and an adapted learning classifier system, have been proposed and preliminarily outlined in (Li and Aickelin, 2004). This paper extends the work in (Li and Aickelin, 2004) and presents a new memetic estimation of distribution algorithm, in which an ant-miner algorithm is embedded as the local search processor to improve the resulting solutions in each generation. Memetic approaches can be thought of as hybridisations between evolutionary techniques and local search methods (e.g. see Krasnogor and Smith, 2005; Krasnogor et al, 2004).

The work presented in this paper is motivated by the goal of attempting to explore how decision support systems can learn to solve new problems. This is, of course, one of the major motivations for recent work on hyper-heuristics (see Burke et al, 2003a; Ross, 2005). Hyper-heuristics can be thought of as "*heuristics to choose heuristics*" (Ross, 2005). Recent work in this area has dealt with a range of methodologies for selecting which heuristics to employ. Examples include simulated annealing (e.g. Dowsland et al, 2006), genetic algorithms (e.g. Ross et al, 2003), choice functions (e.g. Rattadilok et al, 2005), case based reasoning (e.g. Burke et al, 2006) and tabu search (e.g. Kendal and Mohd Hussin, 2003). Indeed, Burke et al (2003b) presents a tabu search hyper-heuristic that is evaluated on a university course timetabling problem and a variant of the nurse rostering problem that is addressed in this paper. The authors use a tabu search mechanism to select "*low level*" heuristics to solve the problem. The emphasis in much of the current body of work on hyper-heuristics is the goal of developing systems that are more readily suitable for different problem solving environments than the current state of the art. No direct comparisons of results are possible, as Burke et al (2003) used randomized data and reported results in terms of robustness across different problems rather than solution quality, but our general aim in this paper is the same. We have developed a methodology that can learn which nurse-rule pairs are most appropriate.

The long-term aim of this research is to model the learning of a human scheduler. Humans can provide high quality solutions, but this is tedious and time consuming. Typically, they construct schedules based on rules learnt during scheduling. Due to human limitations, these rules are typically simple. Hence, our rules will be relatively simple too. Nevertheless, human generated schedules are of high quality due to the ability of the scheduler to switch between the rules, based on the status of the current partial solution. One of the goals of the research presented in this paper is that our proposed algorithm should perform this task.

## 2 The Nurse Rostering Problem

The problem addressed here is concerned with creating schedules for wards containing up to 30 nurses. The data is taken from a major UK hospital. The solutions must adhere to constraints generated by staff contracts. They must also satisfy the requirement to have a certain number of nurses of different grades on each shift. Finally, the solutions have to be seen to be fair to all nurses. Although the hospital uses a planning horizon of five weeks, to reduce complexity, our model employs just one week. However, this does not affect solution quality because historic information, such as weekends worked in previous weeks, is fully incorporated into the penalty costs $p_{ij}$. A particular challenge with this problem lies in the definition of *grade*, which facilitates the substitute of nurses of higher grades to nurses of lower grades but not vice versa.

The day is partitioned into three shifts: two day shifts called 'earlies' and 'lates', and a longer night shift. Until the final scheduling stage, 'earlies' and 'lates' are merged into day shifts. A nurse normally works either days or nights in a given week, and a full week's work usually includes more days than nights. For example, a full-time nurse works 5 days or 4 nights, whereas a part-time nurse works 4 days or 3 nights, 3 days or 3 nights, or 3 days or 2 nights. However, exceptions are possible and some nurses have to work both day- and night-shifts in one week.

As described in (Aickelin and Dowsland, 2000), the problem can be decomposed into three independent stages. The first stage uses a knapsack model to check if there are enough nurses to meet demand. If not, additional nurses are allocated to the ward, so that the second stage will always admit a feasible solution. The second stage is the most difficult and is concerned with the actual allocations of days or nights to be worked by each nurse. The third stage then uses a network flow model to assign those on days to 'earlies' and 'lates'. The first and the third stages are relatively easy to solve (Easton and Mansour, 1999). In this paper, we are only concerned with the highly constrained second stage, i.e. allocating nurses to specific days and / or nights.

The numbers of days or nights to be worked by each nurse defines the set of feasible weekly work patterns for that nurse. These will be referred to as shift patterns in the following. For example, (1111100 0000000) would be a pattern where the nurse works the first 5 days and no nights. Each possible weekly shift pattern for a given nurse can be represented as a 0-1 vector with 14 elements, where the first 7 elements represent the 7 days of the week and the last 7 elements the corresponding 7 nights of the week. A '0' or '1' in the vector denotes a scheduled day or night 'off' or 'on' respectively. For each nurse, there are a limited number of shift patterns available. For instance, a full-time nurse working either 5 days or 4 nights has a total of 21 (i.e. $C_7^5$) feasible day shift patterns and 35 (i.e. $C_7^4$) feasible night shift patterns.

Typically, there are between 20 and 30 nurses per ward, 3 grade-bands, 9 part time options leading to 411 different shift patterns, which are listed in detail in (Aickelin and Dowsland, 2000). Thus, the integer programming formulation below has some 12330 binary variables and 72 constraints. Depending on the nurses' preferences, the recent history of patterns worked, and the overall attractiveness of the pattern, a penalty cost is allocated to each nurse-shift pattern pair. These values were set in close consultation with the hospital and range from 0 (perfect) to 100 (unacceptable), with a bias towards lower values.

The problem has previously been addressed in (Aickelin and Dowsland, 2004; Burke et al, 2003). It can be formulated as an integer linear program as follows.
Target function:

$$\text{Minimize} \sum_{i=1}^{n} \sum_{j \in F(i)}^{m} p_{ij} x_{ij} \ . \tag{1}$$

Subject to:
1. Every nurse works exactly one feasible shift pattern:

$$\sum_{j \in F(i)}^{m} x_{ij} = 1, \forall i \in \{1,...,n\} \ ; \tag{2}$$

2. The demand for nurses is fulfilled for every grade on every day and night:

$$\sum_{j \in F(i)}^{m} \sum_{i=1}^{n} q_{is} a_{jk} x_{ij} \geq R_{ks}, \forall k \in \{1,...,14\} \text{ and } s \in \{1,...,g\} \ . \tag{3}$$

Where
$m$ = number of shift patterns;
$n$ = number of nurses;
$g$ = number of grades;
$x_{ij}$ = 1 if nurse $i$ works shift pattern $j$, 0 otherwise;
$p_{ij}$ = preference cost of nurse $i$ working shift pattern $j$;
$F(i)$ = set of feasible shift patterns for nurse $i$;
$q_{is}$ = 1 if nurse $i$ is of grade $s$ or higher, 0 otherwise;
$a_{jk}$ = 1 if shift pattern $j$ covers day/night $k$ (1–7 are days and 8–14 are nights), 0 otherwise;
$R_{ks}$ = demand of nurses with grade $s$ on day/night $k$;

## 2.1 A Graphic Representation of the Solution Space

The nurse rostering problem can be solved by transforming the original problem into another rule-based problem, whose solution space is represented as a hierarchical and acyclic directed graph (shown in Figure 1). The node $N_{ij} (i \in \{1,...,n\}; j \in \{1,...,r\})$ in the graph denotes that nurse $i$ is assigned by using rule $j$, where $n$ is the number of nurses to be scheduled and $r$ is the number of rules available in the building process. The directed edge (arrow) from node $N_{ij}$ to node $N_{i+1,j'}$ denotes a causal relationship represented by "$N_{i+1,j'}$ following $N_{ij}$", i.e. a rule link from nurse $i$ to nurse $i+1$ indicating nurse $i$ is scheduled by rule $j$ and nurse $i+1$ by rule $j'$. In this graph, a possible solution is represented as a directed path from nurse 1 to nurse $n$ connecting $n$ nodes.

Figure 1 about here

# 3 A Construction Heuristic for the Rule-Based Problem

Our method has been motivated by observing human working patterns. A building heuristic is designed to form a schedule step by step by using a rule set consisting of six heuristic rules. Note that other rules could also be added into the rule set.

## 3.1 Six Building Rules

The first rule, called '*Random*' rule, is used to select a nurse's shift pattern at random. Its purpose is to introduce randomness into the search thus enlarging the search space, and most importantly to ensure that the proposed algorithm has the ability to escape from local optima. This rule mirrors much of a scheduler's creativeness to generate different solutions if required.

The second rule is the '*k-Cheapest*' rule. Disregarding the feasibility of the solution, it randomly selects a shift pattern from a *k*-length list containing patterns with the *k*-cheapest costs $p_{ij}$, in an effort to reduce the schedule cost as much as possible.

The third '*Highest Undercover*' rule is designed to consider only the feasibility of the schedule. It schedules one nurse at a time in a shift pattern that has the highest number of uncovered shifts. For instance, assume a nurse is able to work under two shift patterns: one covers Monday to Friday nights and the other covers Tuesday to Saturday nights. Furthermore, assume that the current requirements for the nights from Monday to Sunday are as follows: (-3, 0, +1, -2, -1, -2, 0), where a negative number represents *undercover* and a positive number signifies *over cover*. The Monday to Friday shift pattern hence has a cover value of 3 as the most negative value it covers is -3, and the Tuesday to Saturday pattern has a value of 2. The '*Highest Undercover*' rule would assign the first pattern to this nurse due to its larger undercover value. Note that, in order to ensure that high-grade nurses are not '*wasted*' covering unnecessarily for nurses of lower grades, for nurses of grade *s*, only the shifts requiring grade *s* nurses are counted as long as there is a single uncovered shift for this grade. If all these are covered, shifts of the next lower grade are considered and once these are filled those of the next lower grade.

The fourth rule, '*Overall Cover*', is very similar to the third rule '*Highest Cover*'. The difference is that the '*Overall Cover*' rule tries to find shift patterns with the largest amount of overall undercover, which is the sum of individual undercover of each shift. Using the above example, the Monday to Friday shift pattern has an overall cover value of 6 as the sum of undercover is -6, while the Tuesday to Saturday pattern would have an overall cover value of 5. The Monday to Friday pattern is therefore assigned.

The fifth '*Contribution-A*' rule is biased towards solution quality but includes some aspects of feasibility by computing an overall score for each feasible pattern for the nurse currently being scheduled. It is designed to take into account the nurses' preferences. It also takes into account some covering constraints in which it gives preference to patterns that cover shifts that have not yet been allocated sufficient nurses to meet their total requirements. In summary, this rule is to go through the entire set of feasible shift patterns for a nurse and assign each one a score. The one with the highest (i.e. best) score is chosen. The score of a shift pattern $S_{ij}$ can be denoted as

$$S_{ij} = \sum_{s=1}^{3} w_s q_{is} (\sum_{k=1}^{14} a_{jk} d_{ks}) + w_4 (100 - p_{ij}), \qquad (4)$$

where $w_s, s \in \{1,2,3\}$, is the weight of covering an uncovered shift of grade $s$, $w_4$ is the weight of the nurse's $p_{ij}$ value for the shift pattern, $a_{jk}$ is 1 if shift pattern $j$ covers day $k$ and 0 otherwise, and $d_{ks}$ is 1 if there are still nurses needed on day $k$ of grade $s$ and 0 otherwise. Note that ($100-p_{ij}$) must be used in the score, as higher $p_{ij}$ values are worse and the maximum for $p_{ij}$ is 100.

The sixth rule, '*Contribution-B*', also uses formula (4) to assign each feasible shift pattern a score, but it uses a different definition of $d_{ks}$. Here $d_{ks}$ equals the actual number of nurses required if there are still nurses needed on day $k$ of grade $s$, 0 otherwise.

### 3.2 Fitness Function

Our nurse rostering problem is complicated by the fact that higher qualified nurses can be substituted for nurses with lower qualifications but not vice versa. Furthermore, the problem has a special day-night structure as most of the nurses are contracted to work either days or nights in one week but not both. These two characteristics mean that the finding and maintaining of feasible solutions in any heuristic search is extremely difficult. Therefore, a penalty function approach is needed while calculating the fitness of completed solutions. Since the chosen encoding automatically satisfies constraint (2) of the integer programming formulation, we can use the following formula to calculate the fitness of solutions (the fitter the solution, the lower its fitness value):

$$\text{Minimize} \quad \sum_{i=1}^{n}\sum_{j=1}^{m} p_{ij} x_{ij} + w_5 \sum_{k=1}^{14}\sum_{s=1}^{p} \max\left(\left[R_{ks} - \sum_{i=1}^{n}\sum_{j=1}^{m} q_{is} a_{jk} x_{ij}\right], 0\right). \quad (5)$$

Note that only undercovering is penalized not overcovering. Hence, the *max* function is used in formula (5). The parameter $w_5$ is the penalty weight used to adjust the penalty that a solution has added to its fitness, and this penalty is proportional to the number of uncovered shifts. For example, consider a solution with an objective function value of 15 that undercovers the Monday day shift by one shift and the Tuesday night shift by two shifts. If the penalty weight was set to 20, the fitness of this solution would be 15 + (1+2)*20 = 75.

## 4 An Estimation of Distribution Algorithm with Intelligent Local Search

An Estimation of Distribution Algorithm (EDA) (Bosman and Thierens, 2000; Larranaga, 2002; Larranaga and Lozano, 2002; Pelikan et al, 2002) is a probabilistic model-building genetic algorithm. However, unlike a traditional genetic algorithm which applies crossover to pairs of selected solutions and then applies mutation to each of the resulting solutions (e.g. see Sastry et al, 2005), EDAs build a probabilistic model based on selected solutions and then sample the model to generate new candidate solutions. This section introduces a memetic EDA, in which a local search procedure conducted by an ant-miner algorithm is integrated in the EDA. Such a procedure is applied to the solutions generated by the model-sampling process of the EDA before they are returned as new candidate solutions.

### 4.1 Probabilistic Model-Building and Model-Sampling by Bayesian Networks

A Bayesian network (Pearl, 1988; Pelikan, 2005) comprises a structure and a set of parameters, where the structure is encoded by a directed acyclic graph with the nodes corresponding to the variables in the modelled data set. The edges in a Bayesian network correspond to conditional dependencies between nodes, and the parameters are represented by a set of conditional probability tables. Bayesian networks are often used to model multinomial data with both discrete and continuous variables by encoding the conditional dependence relationships between the variables contained in the modelled data.

Mathematically, an acyclic Bayesian network encodes a full joint probability distribution by the product

$$p(X) = \prod_{i=1}^{n} p(X_i \mid pa(X_i)), \qquad (6)$$

where $X = (X_1, \ldots, X_n)$ is a vector of all variables in the problem, $pa(X_i)$ is the set of parents of $X_i$ in the network (the set of nodes from which there exists an individual edge to $X_i$), and $p(X_i \mid pa(X_i))$ is the conditional probability of $X_i$ conditioned on its parents $pa(X_i)$. This distribution can be used to generate new instances using the marginal and conditional probabilities.

Learning in a Bayesian network refers to the structure (i.e. topology) of the model, or the parameters (i.e. the values of the conditional probabilities), or both depending on whether the topology of the network is fixed (Mühlenbein and Mahnig, 1999) or not (Pelikan, 2005). The directed graph shown in Figure 1 denotes the solution structure of the problem, which represents, in essence, a fixed nurse-size vector of rules. In this model, the network structure is fixed, and thus learning in the Bayesian network, in our case, refers to parametric learning only. The goal of learning in this situation is to find the parameters of each conditional probability distribution that maximizes the likelihood of the population consisting of a number of promising solutions.

Compared with structural learning, parametric learning for a given structure is simple, because the value of each variable in the population of promising solutions is fully observed. In our proposed network, learning amounts to counting and hence we use the symbol '#' meaning 'the number of' in the following equations. It calculates the conditional probabilities of each possible value for each node given all possible values of its parents. For example, for node $N_{i+1,j'}$ with a parent $N_{ij}$, its conditional probability is computed as

$$P(N_{i+1,j'} \mid N_{ij}) = \frac{P(N_{i+1,j'}, N_{ij})}{P(N_{ij})} = \frac{\#(N_{i+1,j'} = true, N_{ij} = true)}{\#(N_{i+1,j'} = true, N_{ij} = true) + \#(N_{i+1,j'} = false, N_{ij} = true)}. \qquad (7)$$

Note that nodes $N_{1j}$ in the first hierarchical layer have no parents. In this circumstance, their probabilities are computed as

$$P(N_{1j}) = \frac{\#(N_{1j} = true)}{\#(N_{1j} = true) + \#(N_{1j} = false)} = \frac{\#(N_{1j} = true)}{\#Training\ sets}. \qquad (8)$$

To facilitate the understanding of how these probabilities are computed, let us create a simple dummy problem of scheduling three nurses by three rules (shown in Figure 2 below). The scheduling process is repeated 50 times. Each time, rules are randomly used to get a solution, whether it is feasible or not. The value adjacent to each edge represents the total number of times that this edge has been used in the 50 runs. For example, if one of the solutions is obtained by using rule 2 to schedule nurse 1, rule 3 to nurse 2 and rule 1 to nurse 3, then there exists a path of "$N_{12} \rightarrow N_{23} \rightarrow N_{31}$", and the count of edge "$N_{12} \rightarrow N_{23}$" and edge "$N_{23} \rightarrow N_{31}$" are increased by one respectively.

Figure 2 about here

Therefore, we can calculate the (conditional) probabilities of each node according to the above count. For the three nodes that have no parents, their probabilities are calculated as:

$$P(N_{11}) = \frac{10+2+3}{(10+2+3)+(5+11+4)+(7+5+3)} = \frac{15}{50}, \quad P(N_{12}) = \frac{5+11+4}{50} = \frac{20}{50}, \quad P(N_{13}) = \frac{7+5+3}{50}.$$

For all other nodes that have parents, their conditional probabilities are calculated as:

$$P(N_{21}|N_{11}) = \frac{10}{10+2+3} = \frac{10}{15}, \quad P(N_{22}|N_{11}) = \frac{2}{10+2+3} = \frac{2}{15}, \quad P(N_{23}|N_{11}) = \frac{3}{10+2+3} = \frac{3}{15},$$

$$P(N_{21}|N_{12}) = \frac{5}{5+11+4} = \frac{5}{20}, \quad P(N_{22}|N_{12}) = \frac{11}{5+11+4} = \frac{10}{20}, \quad P(N_{23}|N_{12}) = \frac{4}{5+11+4} = \frac{4}{20},$$

$$P(N_{21}|N_{13}) = \frac{7}{7+5+3} = \frac{7}{15}, \quad P(N_{22}|N_{13}) = \frac{5}{7+5+3} = \frac{5}{15}, \quad P(N_{23}|N_{13}) = \frac{3}{7+5+3} = \frac{3}{15},$$

$$P(N_{31}|N_{21}) = \frac{7}{7+9+3} = \frac{7}{19}, \quad P(N_{32}|N_{21}) = \frac{9}{7+9+3} = \frac{9}{19}, \quad P(N_{33}|N_{21}) = \frac{3}{7+9+3} = \frac{3}{19},$$

$$P(N_{31}|N_{22}) = \frac{11}{11+1+5} = \frac{11}{17}, \quad P(N_{32}|N_{22}) = \frac{1}{11+1+5} = \frac{1}{17}, \quad P(N_{33}|N_{22}) = \frac{5}{11+1+5} = \frac{5}{17},$$

$$P(N_{31}|N_{23}) = \frac{10}{10+4+0} = \frac{10}{14}, \quad P(N_{32}|N_{23}) = \frac{4}{10+4+0} = \frac{4}{14}, \quad P(N_{33}|N_{23}) = \frac{0}{10+4+0} = \frac{0}{14}.$$

Once the parameters of the Bayesian network have been learned, new candidate solutions (i.e. new paths) are generated according to the distribution encoded by the learned network (see formula 6). This can be done by applying the method of probabilistic logic sampling (Pelikan, 2005), which stochastically instantiates the Bayesian network, beginning with the root nodes and using the appropriate conditional distribution to extend the instantiation through the network. For our nurse rostering problem, since the first ancestral node in a solution has no parents, it will be chosen from nodes $N_{1j}$ in the first layer according to their probabilities. The next node will be chosen from nodes $N_{ij}$ according to their probabilities conditioned on the previous nodes. This building process is repeated until the last node has been chosen from nodes $N_{nj}$, where $n$ is the number of the nurses. A link from nurse 1 to nurse $n$ is thus created, representing a new possible solution. Since all the probability values are normalized, the roulette-wheel method can be used for rule selection.

### 4.2 A Memetic EDA for Nurse Rostering
Based on an estimation of conditional probabilities, we present a memetic EDA for the nurse rostering problem. It uses techniques from the field of modelling data by Bayesian networks to estimate the joint distribution of promising solutions. The nodes, or variables, in the Bayesian network correspond to the individual nurse-rule pairs by which a schedule will be built step by step.

The general procedure of the proposed memetic EDA is similar to that of a memetic genetic algorithm. The initial population of candidate solutions, or paths, is generated at random according to a uniform distribution over the space of all potential paths. In each generation, a set of better paths is first selected from the current population. Any selection method biased towards better fitness can be used, and in this paper, the traditional roulette-wheel selection (i.e. proportional selection) is applied to sample the obtained selection probabilities. A probabilistic model based on the Bayesian network described in Section 4.1 is then built and the (conditional) probabilities of each node in the Bayesian network are computed. This built probabilistic model is then sampled to generate new paths by using these conditional probability values. Subsequently, an ant-miner algorithm implements local search on each resulting path and thus hopefully produces a set of improved paths or solutions. These newly produced solutions are then added into the old population, replacing some of the old solutions. The process is repeated until stopping conditions are reached. The memetic EDA procedure is outlined in Figure 3.

Figure 3 about here

### 4.3 An Ant-Miner Algorithm for Local Refinement

In nature, ants searching for food are able to find the shortest path between a food source and their nest by exchanging information via pheromones. This substance is laid on the ground when an ant moves, thus marking its path by a pheromone trail. While an isolated ant moves essentially at random, an ant encountering a previously laid trail can detect it and is attracted to it with high probability, thus reinforcing the trail with its own pheromone. The more ants follow a trail, the more attractive that trail becomes for the following ants. The process is thus characterized by a positive feedback loop, in which the probability that an ant will choose a path increases with the number of ants that previously chose the same path.

Ant colony algorithms (Dorigo and Stützle, 2004) are inspired by the behaviour of real ants. In ant colony algorithms, artificial ants are used to search for good quality solutions to the combinatorial optimization problems under considered. Each ant constructs a complete solution by starting with a null solution and adding a solution component at each step until a complete solution is constructed. The selection of this added component is influenced through problem-specific heuristic information as well as (artificial) pheromone trails. After an artificial ant has constructed a feasible solution, the pheromone trails are updated depending on the objective function value of the constructed solution. This update will influence the selection process of the next ants.

Using ant colony algorithms in nurse rostering is a rather new approach. The only related work we have found in the literature is an *in press* paper (Gutjahr and Rauner, 2006), which solves a dynamic regional nurse-scheduling problem in Austria. Our ant-miner algorithm also takes ideas from the ant colony paradigm and has been adapted to a different nurse rostering problem in the UK to improve individual solutions further. It is more like a simplified ant colony algorithm by allowing just one ant in each ant-cycle to speed up the local search process. Each move that an ant takes corresponds to a node (i.e. a nurse-rule pair shown in Figure 1) which has some amount of pheromone indicating how good the choice of that node was in former runs. The amount of pheromone on each node is constantly assessed and updated, intending to reinforce nodes that are used in better solutions. Since the nurse rostering problem that we are addressing in this paper is a minimization problem, once a solution is constructed, some amount of pheromone, which is inversely proportional to the cost of the resulting solution, will be evenly laid on all nodes in the solution. The smaller the solution cost, the greater the amount of pheromone assigned to each node in the solution.

We call a cycle of the ant-miner algorithm a *generation*, after which a path from nurse 1 to nurse $n$ connecting $n$ nodes is constructed and a new solution is thus obtained. Let $\tau_{ij}(t)$ be the intensity of the pheromone trail for node $N_{ij} (i \in \{1,...,n\}; j \in \{1,...,r\})$, where $n$ is the number of nurses and $r$ is the number of rules used to schedule nurses, at generation $t$. The trail intensity of node $N_{ij}$ at the next generation ($t$+1) is updated according to the following formula

$$\tau_{ij}(t+1) = \rho \times \tau_{ij}(t) + \Delta\tau_{ij}, \qquad (9)$$

where $\rho$ is a coefficient such that (1-$\rho$) represents the evaporation of each trail between two generations, and the value of $\rho$ must be smaller than 1 to avoid unlimited accumulation of pheromone trails. $\Delta\tau_{ij}$ is the amount of pheromone laid on node $N_{ij}$, which is defined as

$$\Delta \tau_{ij} = \begin{cases} \dfrac{Q \times D}{c_t}, & \text{if node } N_{ij} \text{ is used to build the current solution} \\ 0, & \text{otherwise} \end{cases} \quad (10)$$

and

$$D = \begin{cases} d, & \text{if the current solution is the best solution found so far} \\ 1, & \text{otherwise} \end{cases} \quad (11)$$

where $Q$ is a constant, $c_t$ is the solution cost obtained at the $t$-th generation and $d$ is also a constant representing the additional reinforcement times if the resulting solution is also the best solution found so far.

As a local search processor, the ant-miner algorithm starts the search from individual solutions already produced by the main EDA. In each solution's refinement process, high pheromone levels must be deposited on the initial trail in order for later miners to follow the trail with high probability. Thus, we initialize the trail intensities of node $N_{ij}$ for each ant-miner algorithm started from each generation of the EDA as

$$\tau_{ij}(0) = \begin{cases} B_1, & \text{if node } N_{ij} \text{ is used in the initial solution for an ant} \\ B_2, & \text{otherwise} \end{cases} \quad (12)$$

where $B_1$ and $B_2$ are constants, satisfying $B_1 > B_2$.

In determining which nodes should be used in building a schedule, nodes with higher trail intensity are more desirable. Thus, we can define the probability of choosing node $N_{ij}$ at generation $t$ as

$$p_{ij}(t) = \frac{\tau_{ij}(t)}{\sum_{j=1}^{r} \tau_{ij}(t)}, \forall i \in \{1,2,\ldots,n\}. \quad (13)$$

Given the definitions above, we design an ant-miner algorithm to implement local refinement starting from individual resulting solutions in each generation of the memetic EDA (described in Section 4.2). The general structure of the proposed algorithm is outlined in Figure 4.

Figure 4 about here

To explain how the pheromone trails are laid and reinforced, we will give a simple example of scheduling three nurses by four rules. In the trail intensity updating formulae (10), (11) and (12), we set constant $Q = 4$, $d = 2$, coefficient $\rho = 0.5$, $B_1 = 5$ and $B_2 = 1$. The initial solution $S_0$, with an associated cost of 5, is generated by using "rule 1 for nurse 1, rule 4 for nurse 2 and rule 3 for nurse 3" denoted as the node sequence "$N_{11} \to N_{24} \to N_{33}$". The next solution, with a cost of 4, is generated by using "rule 4 for nurse 1, rule 2 for nurse 2 and rule 3 for nurse 3" denoted as "$N_{14} \to N_{22} \to N_{33}$". Thus, the trail intensity matrix (13) at each generation is updated as follows:

$$\begin{array}{cc} \text{Gen. 0} & \text{Gen. 1} \end{array}$$

$$\begin{pmatrix} 5 & 1 & 1 & 1 \\ 1 & 1 & 1 & 5 \\ 1 & 1 & 5 & 1 \end{pmatrix} \Rightarrow \begin{pmatrix} 5/2 & 1/2 & 1/2 & 1/2 + (4 \times 2)/4 \\ 1/2 & 5/2 + (4 \times 2)/4 & 1/2 & 5/2 \\ 1/2 & 1/2 & 5/2 + (4 \times 2)/4 & 1/2 \end{pmatrix} = \begin{pmatrix} 2.5 & 0.5 & 0.5 & 2.5 \\ 0.5 & 2.5 & 0.5 & 2.5 \\ 0.5 & 0.5 & 4.5 & 0.5 \end{pmatrix}$$

# 5 Computational Results

In this section, we present the results of extensive experiments on a nurse rostering problem with 52 real data instances collected from a major UK hospital and we compare them to results of the same data instances found previously by other algorithms.

## 5.1 Details of Algorithms

Table 1 lists detailed computational results of various approaches over 52 instances. The results listed in Table 1 are based on multiple runs with different random seeds and the last row contains the mean values of all columns (N/A indicates no feasible solution was found):

- IP: optimal solutions found with a commercial IP software package (called XPRESS MP) (Dowsland and Thompson, 2000);
- GA: best result out of 20 runs from a parallel GA with multiple sub-populations and intelligent parameter adaptation (Aickelin and White, 2004);
- Rd-1: best result of 20,000 iterations of a random search, i.e. only the first random rule is in use;
- Rd-2: best results of 20,000 iterations of a search with random rule-selection, i.e. using six rules but every rule has an equal opportunity to be chosen all the time for all nurses;
- Cost: best result out of 20 runs;
- Inf: number of runs terminating with the best solution being infeasible;
- #: number of runs terminating with the best solution being optimal or equal to the best known;
- <3: number of runs terminating with the best solution being within three cost units of the optimum. The value of three units was chosen as it corresponds to the penalty cost of violating the least important level of requests in the original formulation. Thus, these solutions are still acceptable to the hospital.

For all data instances, we used the following set of fixed parameters to implement our experiments. These parameters are based on our experience and intuition and thus are not necessarily the best for each instance. We have kept them the same for consistency at this stage.

- Number of generations: 200 for the EDA and 5 for each ant-miner algorithm;
- For the '$k$-Cheapest' rule, $k = 5$;
- Weight set in formula (4): $w_1 = 1$, $w_2 = 1$, $w_3 = 8$ and $w_4 = 2$;
- Penalty weight in fitness function (5): $w_5 = 200$;
- The number of solutions kept in each generation = 140 (EDA only), in which 100 solutions are selected by the roulette wheel method and remaining ones are the best 40 solutions in the previous generation;
- The trail intensities updating in formulae (9), (10) and (11): $\rho = 0.97$, $Q = 100$ and $d = 2$;
- The initial trail densities in formula (12): $B_1 = 10$ and $B_2 = 1$;
- Number of runs = 20.

Our proposed algorithm was coded in Java 2, and all instances were run on the same Pentium 4 2.0GHz PC with 512MB RAM under the Windows XP operating system. The execution time per run per data instance is approximately half a minute for the EDA without local refinement and 2-3 minutes for the EDA with local refinement. With regard to the IP and the parallel GA, they were run on a different Pentium III PC. This took in excess of 24 hours (ILP) and approximately half a minute (single parallel GA run using C) respectively. However, an accurate comparison in terms of CPU time among the three algorithms is difficult due to the different environments that were in use. We still conclude that our algorithm is much faster than the ILP and probably a bit slower than the GA, particularly when we implement local refinement in the EDA. However, if we compare the EDA with the GA based on the 'number of evaluations' (e.g. population size times number of generations), then they are approximately the same.

### 5.2 Analysis of Results

First, let us discuss the results in Table 1. Comparing the computational results on various test instances, one can see that using the random rule alone does not yield a single feasible solution, as the 'Rd.1' column shows. This underlines the problem difficulty. In addition, without learning, the results of randomly selecting one of the six rules at each move are much weaker, as the 'Rd.2' column shows. Thus, it is not simply enough to use the six rules to build schedules. With regard to the results of EDAs with and without local search, in general, the results found by the EDA without local search rival those found by a complex multi-population GA (with features of competing sub-populations and self-learning for good parameters), and the results by the EDA with local refinement are much better. Particularly impressive is the fact that, for both algorithms, a feasible solution is found in 100% of cases. Note that independently of the algorithm used, some data instances (i.e. 29, 31, 32, 50, 51 and 52) are harder to solve than others due to a shortage of nurses in some weeks.

Table 1 about here

Figures 5 and 6 show the results of the memetic EDA and the multi-population GA graphically. The bars above the *x*-axis represent solution quality: the black bars show the number of optimal solutions found (i.e. the value of '#' in Table 1), the grey acceptable solutions (i.e. the value of '<3' in Table 1). The margin of '3' is chosen because under the current hospital rules this represents the lowest level of a penalty for a single unfulfilled request and is still regarded as 'high quality'. The bars below the *x*-axis represent the number of times the algorithm failed to find a feasible solution (i.e. the value of 'Inf' in Table 1). Hence, the shorter the bar is below the *x*-axis and the longer above, the better the algorithm's performance. Note that 'missing' bars mean that feasible, but not optimal solutions were found.

Figure 5 shows that for the memetic EDA, 41 out of 52 data instances are solved to or near to optimality (within three cost units), and feasible solutions are always found for all data sets. For the genetic algorithm's performance shown in Figure 6, the results are similar: 42 data sets are solved well. However, many solutions are infeasible and, for two instances, not a single feasible solution had been identified. All algorithms except the IP one have difficulties in solving the data instances with nurse shortages, but the memetic EDA performs much better.

Figure 5 about here

Figure 6 about here

The behaviour of an individual run of the memetic EDA is as expected. By using a sample data instance 04, Figure 7 depicts its searching process. In this figure, the *x*-axis represents the number of generations and the *y*-axis represents the best solution cost found in each generation consisting of a number of candidate solutions. As shown in Figure 7, the optimal solution, with a cost of 17, is achieved at the generation of 57. Although the actual values may differ among various instances, the characteristic shapes of the curves are similar for all seeds and data instances.

Figure 7 about here

Figure 8 gives the optimal or best-known solutions found by an IP software package, and compares the performance of different GAs (Aickelin and Dowsland, 2000; Aickelin and Dowsland, 2004) with two versions of our EDAs (i.e. with and without local search). These different versions of GAs are briefly explained as follows: The 'Basic GA' is a GA with standard genetic operators; the 'Adapt GA' is the same as the basic GA, but it also tries to self-learn good parameters during the runtime; the 'Multipop GA' is the same as the 'Adapt GA', but it also features competing sub-populations; the 'Hillclimb' GA is the same as the 'Multipop GA', but it also includes a local search in the form of a hill-climber around the current best solution. The comparison results are encouraging: within similar computational time, most complex GAs are outperformed in terms of feasibility, average and best results. Only the 'Hillclimb GA' performs slightly better in terms of best performance.

Figure 8 about here

## 6 Conclusions and Future Work

This paper presents a memetic EDA for nurse rostering, in which a local search procedure conducted by an ant-miner algorithm is embedded, to improve the resulting solutions in each generation of the EDA. Unlike most existing approaches, the new approach has the ability to build schedules using flexible, rather than fixed rules. Experimental results from real-world nurse rostering problems demonstrate that the proposed approach performs better than most existing approaches. Moreover, our approach is not "hard coded" to certain instances. It has been designed with the goal of being able to learn about new problem solving situations in mind. This emphasis is complementary to recent work on hyper-heuristics (Burke et al, 2003a; Ross, 2005).

Although this work is presented in terms of nurse rostering, it is suggested that the method for explicit learning in our approach could be applied to many other scheduling problems where the schedules will be built systematically according to specific rules. It is also hoped that this research would shed light on the significant issue of how to include human-like learning into scheduling algorithms. It may, therefore, be of interest to practitioners and researchers in the areas of scheduling and evolutionary computation.

Our future work will investigate the identification and extraction of good features in the solutions. Such features can be used more explicitly to solve the problem and hence reduce the search space. Furthermore, they will help the human scheduler to understand the problem better and actually learn from the search process for future rescheduling. Another direction for future research is to study how the order in which the nurses appear in the directed graph might affect solution quality. This paper applies a fixed order throughout the search and, potentially, a variable order would make it possible to improve the solution quality even further.

**Acknowledgements**

The work was funded by the UK's Engineering and Physical Sciences Research Council (EPSRC), under grants GR/S70197/01 and GR/S31550/01.

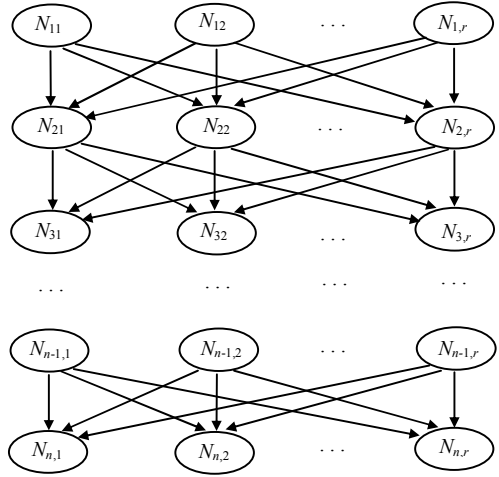

Figure 1: A graphic representation for rule-based nurse rostering

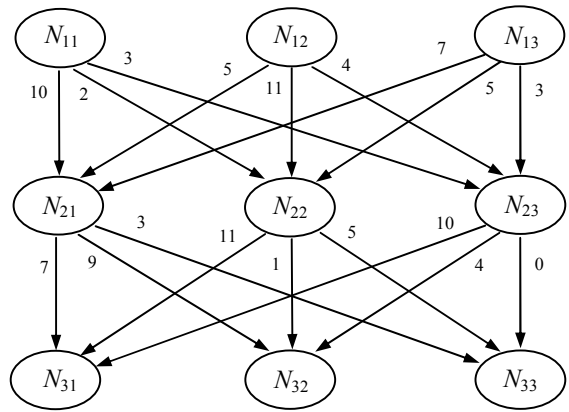

Figure 2: A dummy problem with a size of 3-nurse and 3-rule

```
The memetic EDA for nurse rostering ( )
{
    t'=0;
    Generate an initial population P(0);
    While (stopping condition not reached) {
        Select a set of promising paths S(t') from P(t');
        Build probabilistic model M(t') for S(t');
        Sample M(t') to generate a set of new paths P₁(t');
        Improve each member in P₁(t') by an ant-miner algorithm and
          thus obtain a set of new paths P₂(t');
        Add P₂(t') into P(t');
        t=t+1;
    }
    Return the best path;
}
```

Figure 3: The pseudocode of the memetic EDA

```
The ant-miner algorithm for local refinement ( )
{
    t=0;
    c_best= c_0;
    for (i=0; i<Number of nurses; i++)
        for (j=0; j<Number of rules; j++) {
            If (node N_ij is used in the initial solution) τ_ij(0)=B_1;
            Else τ_ij(0)=B_2;
        }
    While (t< Max. number of generations) {
        t = t+1;
        /* Build a solution according to the intensity of trails */
        for (i=0; i<Number of nurses; i++)
            Use formula (13) to select one rule for nurse i;
        Compute c_t;  /* c_t is the cost of the resulting solution */
        If (c_t ≤ c_best) c_best=c_t;
        Use formula (12) to update τ_ij(t) on all nodes
    }
    Return the best solution cost c_best;
}
```

Figure 4: The pseudo code of the ant-miner algorithm

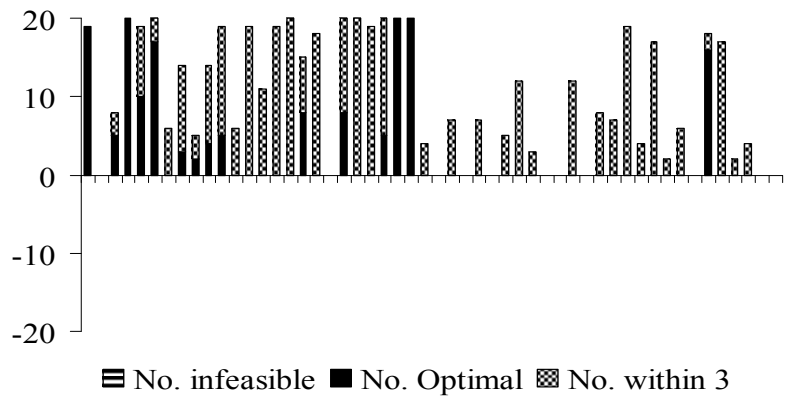

▤ No. infeasible  ■ No. Optimal  ▩ No. within 3

Figure 5: Results of the memetic EDA over 52 instances

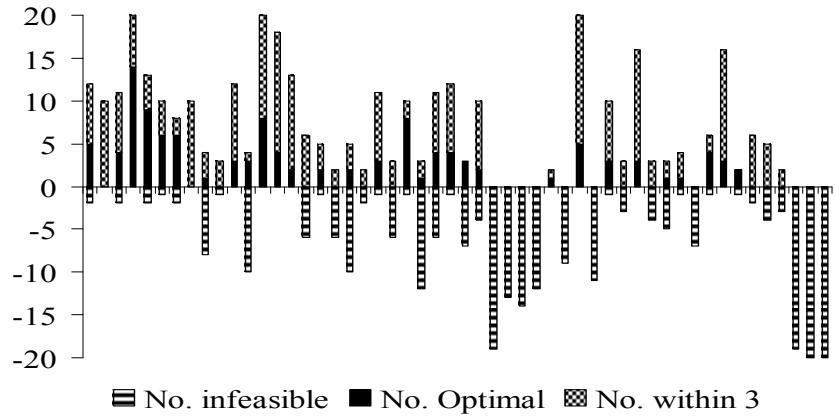

Figure 6: Results of the multi-population GA over 52 instances

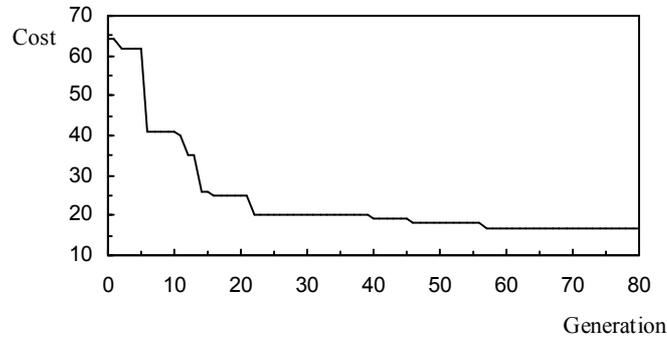

Figure 7: Sample run of the memetic EDA (for instance 04)

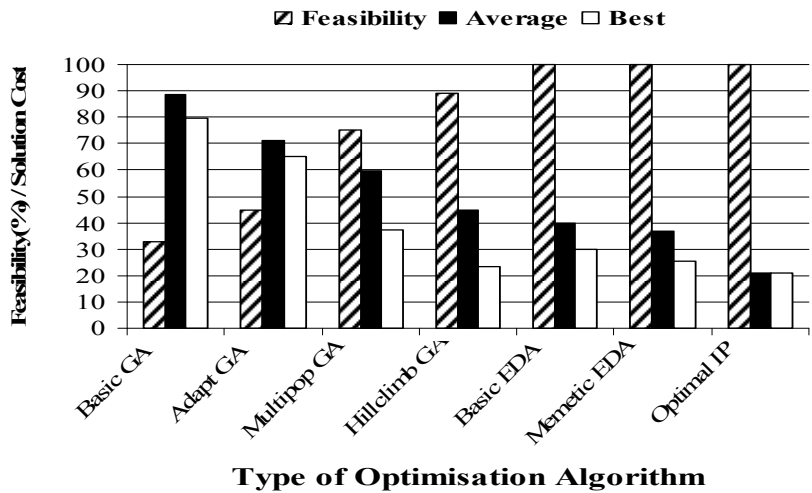

Figure 8: Summary results of various optimization algorithms

| Set | IP | GA | Rd-1 | Rd-2 | EDA without ant-miner | | | | EDA with ant-miner | | | |
|---|---|---|---|---|---|---|---|---|---|---|---|---|
| | | | | | Cost | Inf | # | <3 | Cost | Inf | # | <3 |
| 01 | 8 | 8 | N/A | 27 | 8 | 0 | 19 | 20 | 8 | 0 | 19 | 19 |
| 02 | 49 | 50 | N/A | 85 | 56 | 0 | 0 | 0 | 55 | 0 | 0 | 0 |
| 03 | 50 | 50 | N/A | 97 | 50 | 0 | 2 | 5 | 50 | 0 | 5 | 8 |
| 04 | 17 | 17 | N/A | 23 | 17 | 0 | 20 | 20 | 17 | 0 | 20 | 20 |
| 05 | 11 | 11 | N/A | 51 | 11 | 0 | 8 | 16 | 11 | 0 | 10 | 19 |
| 06 | 2 | 2 | N/A | 51 | 2 | 0 | 17 | 17 | 2 | 0 | 17 | 20 |
| 07 | 11 | 11 | N/A | 80 | 14 | 0 | 0 | 3 | 14 | 0 | 0 | 6 |
| 08 | 14 | 15 | N/A | 62 | 15 | 0 | 0 | 11 | 14 | 0 | 3 | 14 |
| 09 | 3 | 3 | N/A | 44 | 14 | 0 | 0 | 0 | 3 | 0 | 2 | 5 |
| 10 | 2 | 4 | N/A | 12 | 2 | 0 | 2 | 10 | 2 | 0 | 4 | 14 |
| 11 | 2 | 2 | N/A | 12 | 2 | 0 | 2 | 20 | 2 | 0 | 5 | 19 |
| 12 | 2 | 2 | N/A | 47 | 3 | 0 | 0 | 2 | 3 | 0 | 0 | 6 |
| 13 | 2 | 2 | N/A | 17 | 3 | 0 | 0 | 20 | 3 | 0 | 0 | 19 |
| 14 | 3 | 3 | N/A | 102 | 4 | 0 | 0 | 7 | 4 | 0 | 0 | 11 |
| 15 | 3 | 3 | N/A | 9 | 4 | 0 | 0 | 20 | 4 | 0 | 0 | 19 |
| 16 | 37 | 38 | N/A | 55 | 38 | 0 | 0 | 20 | 38 | 0 | 0 | 20 |
| 17 | 9 | 9 | N/A | 146 | 9 | 0 | 4 | 11 | 9 | 0 | 8 | 15 |
| 18 | 18 | 19 | N/A | 73 | 19 | 0 | 0 | 20 | 19 | 0 | 0 | 18 |
| 19 | 1 | 1 | N/A | 135 | 10 | 0 | 0 | 0 | 9 | 0 | 0 | 0 |
| 20 | 7 | 8 | N/A | 53 | 7 | 0 | 5 | 19 | 7 | 0 | 8 | 20 |
| 21 | 0 | 0 | N/A | 19 | 1 | 0 | 0 | 20 | 1 | 0 | 0 | 20 |
| 22 | 25 | 26 | N/A | 56 | 26 | 0 | 0 | 15 | 26 | 0 | 0 | 19 |
| 23 | 0 | 0 | N/A | 119 | 1 | 0 | 0 | 20 | 0 | 0 | 5 | 20 |
| 24 | 1 | 1 | N/A | 4 | 1 | 0 | 20 | 20 | 1 | 0 | 20 | 20 |
| 25 | 0 | 0 | N/A | 3 | 0 | 0 | 18 | 20 | 0 | 0 | 20 | 20 |
| 26 | 48 | 48 | N/A | 222 | 52 | 0 | 0 | 1 | 51 | 0 | 0 | 4 |
| 27 | 2 | 2 | N/A | 158 | 28 | 0 | 0 | 0 | 18 | 0 | 0 | 0 |
| 28 | 63 | 63 | N/A | 88 | 65 | 0 | 0 | 3 | 65 | 0 | 0 | 7 |
| 29 | 15 | 141 | N/A | 31 | 109 | 0 | 0 | 0 | 23 | 0 | 0 | 0 |
| 30 | 35 | 42 | N/A | 180 | 38 | 0 | 0 | 3 | 38 | 0 | 0 | 7 |
| 31 | 62 | 166 | N/A | 253 | 159 | 0 | 0 | 0 | 111 | 0 | 0 | 0 |
| 32 | 40 | 99 | N/A | 102 | 43 | 0 | 0 | 4 | 41 | 0 | 0 | 5 |
| 33 | 10 | 10 | N/A | 30 | 11 | 0 | 0 | 8 | 11 | 0 | 0 | 12 |
| 34 | 38 | 48 | N/A | 95 | 41 | 0 | 0 | 2 | 41 | 0 | 0 | 3 |
| 35 | 35 | 35 | N/A | 118 | 46 | 0 | 0 | 0 | 43 | 0 | 0 | 0 |
| 36 | 32 | 41 | N/A | 130 | 45 | 0 | 0 | 0 | 44 | 0 | 0 | 0 |
| 37 | 5 | 5 | N/A | 28 | 7 | 0 | 0 | 7 | 7 | 0 | 0 | 12 |
| 38 | 13 | 14 | N/A | 130 | 25 | 0 | 0 | 0 | 23 | 0 | 0 | 0 |
| 39 | 5 | 5 | N/A | 44 | 8 | 0 | 0 | 3 | 8 | 0 | 0 | 8 |
| 40 | 7 | 8 | N/A | 51 | 8 | 0 | 0 | 10 | 8 | 0 | 0 | 7 |
| 41 | 54 | 54 | N/A | 87 | 55 | 0 | 0 | 15 | 55 | 0 | 0 | 19 |
| 42 | 38 | 38 | N/A | 188 | 41 | 0 | 0 | 1 | 41 | 0 | 0 | 4 |
| 43 | 22 | 39 | N/A | 86 | 23 | 0 | 0 | 13 | 23 | 0 | 0 | 17 |
| 44 | 19 | 19 | N/A | 70 | 24 | 0 | 0 | 0 | 22 | 0 | 0 | 2 |
| 45 | 3 | 3 | N/A | 34 | 6 | 0 | 0 | 2 | 6 | 0 | 0 | 6 |
| 46 | 3 | 3 | N/A | 196 | 7 | 0 | 0 | 0 | 7 | 0 | 0 | 0 |
| 47 | 3 | 4 | N/A | 11 | 3 | 0 | 13 | 20 | 3 | 0 | 16 | 18 |
| 48 | 4 | 6 | N/A | 35 | 5 | 0 | 0 | 10 | 5 | 0 | 0 | 17 |
| 49 | 27 | 30 | N/A | 69 | 30 | 0 | 0 | 2 | 30 | 0 | 0 | 2 |
| 50 | 107 | 211 | N/A | 162 | 109 | 0 | 0 | 1 | 109 | 0 | 0 | 4 |

| 51 | 74 | N/A | N/A | 197 | 171 | 0 | 0 | 0 | 107 | 0 | 0 | 0 |
| 52 | 58 | N/A | N/A | 135 | 67 | 0 | 0 | 0 | 66 | 0 | 0 | 0 |
| Av | 21.1 | 37.1 | N/A | 82.9 | 29.7 | 0 | 2.5 | 8.5 | 25.2 | 0 | 3.1 | 10.1 |

Table 1: Comparison of results by various approaches over 52 instances